\documentclass{article} 
\usepackage{arxiv,times}
\usepackage{comment}
\usepackage{algorithm}
\usepackage{algorithmic}
\usepackage{tabularx}
\usepackage{array}
\usepackage{booktabs}
\usepackage{enumitem}

\usepackage{amsmath,amsfonts,bm}

\def\eqref#1{equation~\ref{#1}}

\def\1{\bm{1}}

\DeclareMathAlphabet{\mathsfit}{\encodingdefault}{\sfdefault}{m}{sl}
\SetMathAlphabet{\mathsfit}{bold}{\encodingdefault}{\sfdefault}{bx}{n}

\usepackage{amsmath,amssymb,amsfonts,amsthm} 
\usepackage{graphicx}                      
\usepackage{multirow}                      
\usepackage{url}                            
\usepackage{hyperref}
\usepackage{url}

\usepackage{multirow}
\usepackage{graphicx}

\title{AbGaze: Attentive Geometric Representation Learning for End-to-End Antibody Design}

\author{
Jiashuo Wang\(^{1,2,*}\),
Siqi Fan\(^{1,*}\),
Yizhen Luo\(^{1,2}\),
Zaiqing Nie\(^{1,3,\dagger}\)\thanks{
\(^*\)Equal contribution.
\(^\dagger\)Corresponding author.
\(^1\)Institute for AI Industry Research (AIR), Tsinghua University;
\(^2\)Department of Computer Science and Technology, Tsinghua University;
\(^3\)PharMolix Inc.
Correspondence to: Jiashuo Wang \textless wjs25@mails.tsinghua.edu.cn\textgreater, Zaiqing Nie \textless zaiqing@air.tsinghua.edu.cn\textgreater.
\newline\newline
\textit{Preprint. Sep 29, 2026.}
}
}

\begin{document}

\maketitle
\begin{abstract}
Computational antibody design requires representations that capture the geometric patterns underlying antigen--antibody interactions, yet existing approaches often rely on scalar distances or surface-intrinsic features, leaving cross-molecular geometry largely implicit. We present \textbf{AbGaze}, an end-to-end antibody design framework based on attentive geometric representation learning, which encodes distance, spatial direction, and surface-normal orientation of antigen surfaces relative to antibody-residue local frames, and adaptively aggregates these geometric interactions according to their interfacial context. The learned interaction representation is shared across multi-CDR co-design, complex structure prediction, and affinity optimization, with local-frame geometric supervision further constraining the representation. AbGaze outperforms prior methods across all three tasks: relative to the second-best method, it improves amino-acid recovery by 7.1\% and reduces structural error by 14.9\% on average over the six CDRs, improves interface docking quality (DockQ) by 6.6\%, and raises the affinity improvement rate (IMP) by 32.5\%.

\end{abstract}

\section{Introduction}
Antibodies have been widely applied in disease therapy, diagnostics, and biomedical research due to their specific molecular-recognition capabilities~\citep{chan2025fifty}. Antibodies recognize and bind antigenic epitopes through the complementarity-determining regions (CDRs) within their variable domains, while the high sequence and conformational diversity of CDRs enables diverse antigen--antibody binding modes~\citep{selaculang2013recognition,north2011cdr,weitzner2015h3}. Therefore, computational modeling of antibodies conditioned on antigen structures requires accurate characterization of antigen--antibody interactions, where modeling the local spatial relationships at the interface is central to the problem.

Existing methods have evolved from step-by-step pipelines toward end-to-end generation~\citep{jin2022refinegnn}. Early approaches decomposed antibody modeling into sequential stages such as complex structure prediction, CDR design, and side-chain assembly, where intermediate errors could propagate across the pipeline~\citep{jin2022antibody,kong2022conditional,luo2022antigen,sircar2010snugdock}. Recent end-to-end approaches instead jointly model antibody sequences and structures conditioned on antigen information, enabling tighter sequence--structure co-generation within a unified framework~\citep{kong2023end,wang2025iggm,tan2025dyab,wang2026abflow}. In parallel, broader studies of protein interactions have explored interface representations learned from the geometric and chemical features of molecular surfaces~\citep{sverrisson2021dmasif}. Methods such as MaSIF learn interaction fingerprints that capture patterns associated with molecular recognition and surface complementarity, applying them to interaction-site identification and binder design~\citep{gainza2020deciphering,gainza2023masifseed}. These studies demonstrate that local geometric and chemical patterns at molecular interfaces provide informative cues for interaction modeling.

Despite these advances, existing interface representations can capture interaction patterns at the molecular-surface or atomic level, but provide limited explicit characterization of the cross-molecular spatial relationships that depend on local conformations at antigen--antibody interfaces. This challenge is closely related to a problem in protein--protein interaction modeling: how to represent the local geometry of intermolecular contacts. Intermolecular binding is not determined by spatial proximity alone; the distance, direction, and angle to the surface normal at which a molecule approaches the opposing surface all affect local packing and complementarity~\citep{kuroda2016complementarity}, while the contribution of contacts can also depend on their surrounding interface context~\citep{lawrence1993shape,mccoy1997electrostatic,yang2003cooperative}. Accordingly, interaction modeling should explicitly characterize the relative spatial relationships between a paratope conformation and the antigen surface. An effective interaction representation should therefore capture three aspects: spatial proximity between local structures, their relative orientations, and the dependence of local interaction contributions on the surrounding interface context. Distance-based representations capture spatial proximity but reduce each contact to a scalar, making configurations with similar distances but different directions indistinguishable; representations based on surface geometry and chemical features capture local surface patterns, but are not expressed relative to antibody conformations. Because these interactions are defined relative to local antibody conformations, antibody-residue local frames also provide a natural basis for structural supervision: frame-aligned point error (FAPE)~\citep{jumper2021highly} and dihedral-angle constraints couple atoms with residue frames, supplying global structural supervision expressed in the same local coordinates. Given the large space of CDR conformations and binding configurations and the limited number of experimentally resolved antibody--antigen complexes, requiring a model to rediscover these interaction regularities from data can increase the burden on representation learning.

Motivated by these observations, we propose \textbf{AbGaze}, a unified framework for antigen--antibody interaction modeling based on orientation-aware geometry and adaptive attention. Specifically, AbGaze constructs a local interface representation using antibody-residue coordinate frames to explicitly encode distance, spatial direction, and surface-normal information. On top of this representation, an adaptive atom--surface attention scheme, conditioned on interface geometry and fused with contextual node states, learns key local interactions and directs coordinate updates toward the attended surface region, optimized under FAPE and dihedral-angle supervision. This orientation-aware representation acts as a binding prior that unifies multi-CDR co-design, complex structure prediction, and affinity optimization under a single masked-prediction framework, where stochastic masked decoding further yields multiple plausible sequence--structure realizations.

To summarize, our main contributions are as follows:

(i) We propose an orientation-aware local interface representation using antibody-residue coordinate frames to explicitly capture relative distance, spatial direction, and surface normals of the antigen surface.

(ii) We design an adaptive atom--surface attention mechanism conditioned on interface geometry to direct spatial updates and enable geometrically constrained interaction learning under FAPE and dihedral supervision.

(iii) We establish a binding prior that unifies multi-CDR co-design, structure prediction, and affinity optimization within a single masked-prediction framework, enabling multi-candidate sequence--structure decoding.

\begin{figure*}[h]
    \centering
    \includegraphics[width=\textwidth]{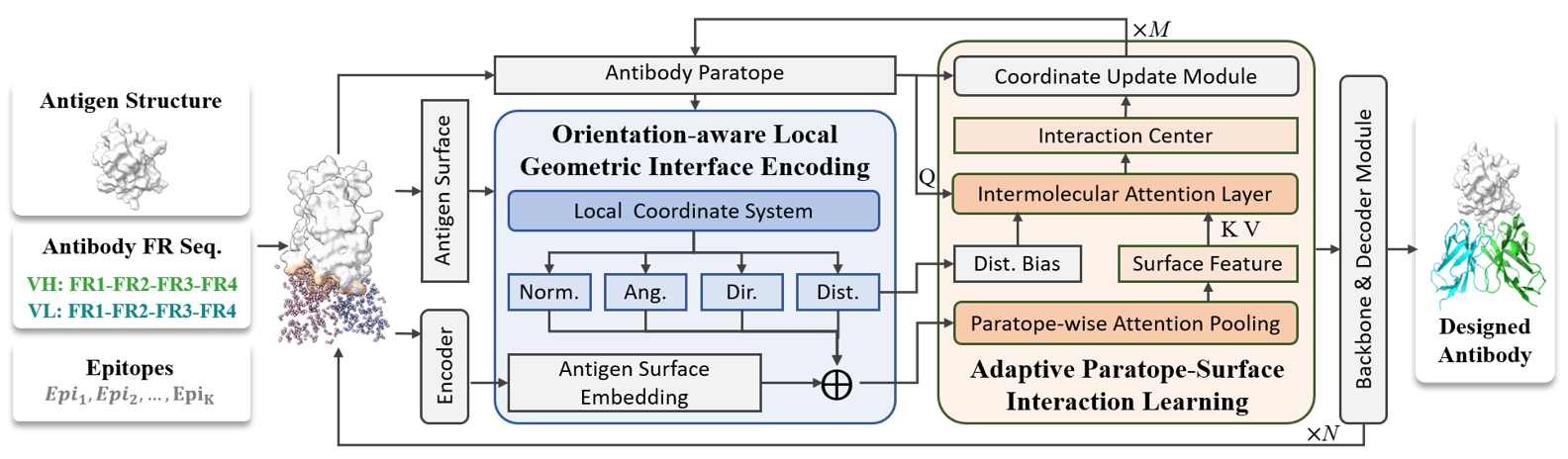}
    \caption{Overview of AbGaze. We propose AbGaze, a unified framework for antigen--antibody interaction modeling based on orientation-aware geometry and adaptive attention. The framework iteratively updates the antibody conformation by explicitly encoding the relative spatial relationships between the paratope local frames and the antigen surface, while an adaptive attention mechanism learns contextual interactions and guides structural refinement.}
    \label{fig:overview}
\end{figure*}

\section{Related Work}
\subsection{Antigen-Conditioned Antibody Modeling}

Antigen-conditioned antibody modeling has evolved from localized CDR design toward end-to-end sequence--structure co-design. Early landmark approaches, including HERN, MEAN, and DiffAb, investigated CDR generation by coupling sequence updates with backbone coordinates~\citep{jin2022antibody,kong2022conditional,luo2022antigen}. Subsequent end-to-end architectures, such as dyMEAN, established scalable full-atom graph dynamics and structural refinement pipelines~\citep{kong2023end}, AbDiffuser paired full-atom diffusion generation with experimental antibody validation~\citep{martinkus2023abdiffuser}, and IgGM broadened generative modeling across diverse functional antibody and nanobody design regimes~\citep{wang2025iggm}. More recently, flow-matching formulations have extended multi-state modeling: dyAb accommodates conformational transitions in target antigens~\citep{tan2025dyab}, and AbFlow leverages surface-guided interaction dynamics for end-to-end design~\citep{wang2026abflow}, alongside geometry-aware frameworks like GeoGAD~\citep{geogad2026}.

\subsection{Molecular Interface Representation}

Characterizing molecular interaction interfaces requires capturing spatial proximity, local orientation, and surface complementarity. Surface-centric methods such as MaSIF and MaSIF-seed showed that geometric fingerprints and intrinsic surface curvature are informative for binding specificity~\citep{gainza2020deciphering,gainza2023masifseed}. This paradigm has been extended to pairwise point-cloud matching in PInet~\citep{dai2021pinet}, structural environment filtering in PeSTo~\citep{krapp2023pesto}, and target-conditioned binder synthesis in ProBID-Net~\citep{chen2024probid}. 

However, integrating such fine-grained surface topography into generative antibody architectures remains non-trivial. Existing generative models typically face a dichotomy in interface representation: they either operate at the residue level via local coordinate frames~\citep{luo2022antigen,geogad2026}, thereby coarse-graining microscopic atomic contact geometry, or include atom-to-surface displacement vectors without decomposing them into residue-local reference frames~\citep{wang2026abflow}. In contrast, AbGaze bridges fine-grained surface topography with frame-aligned local representations. By directly decomposing both antibody-frame approach directions and surface-normal incidence relationships into each residue's local frame, AbGaze aggregates these geometric features through a dedicated two-level surface-attention module for interaction learning and geometric supervision.

\section{Methodology}

\subsection{Overview and Problem Setting}
\label{sec:overview}

We represent an antigen--antibody complex as a residue graph~\citep{jing2021gvp} with amino-acid
types and full-atom coordinates. Context and interface edges are constructed
within and between the two molecules, respectively. For each epitope residue,
we construct an MSMS molecular surface \citep{sanner1996reduced} with vertex
normals, keeping vertices within $10\,\text{\AA}$ of any epitope atom.

Given the antigen sequence and structure and an antibody with masked target
regions, AbGaze jointly generates antibody sequences and full-atom structures
using a shared interface representation and atom--surface interaction
learning. We adopt the general graph-based modeling and decoding framework
of \citet{kong2023end}.

\subsection{Orientation-Aware Local Interface Representation}
\label{sec:representation}

\begin{figure*}[h]
    \centering
    \includegraphics[width=\textwidth]{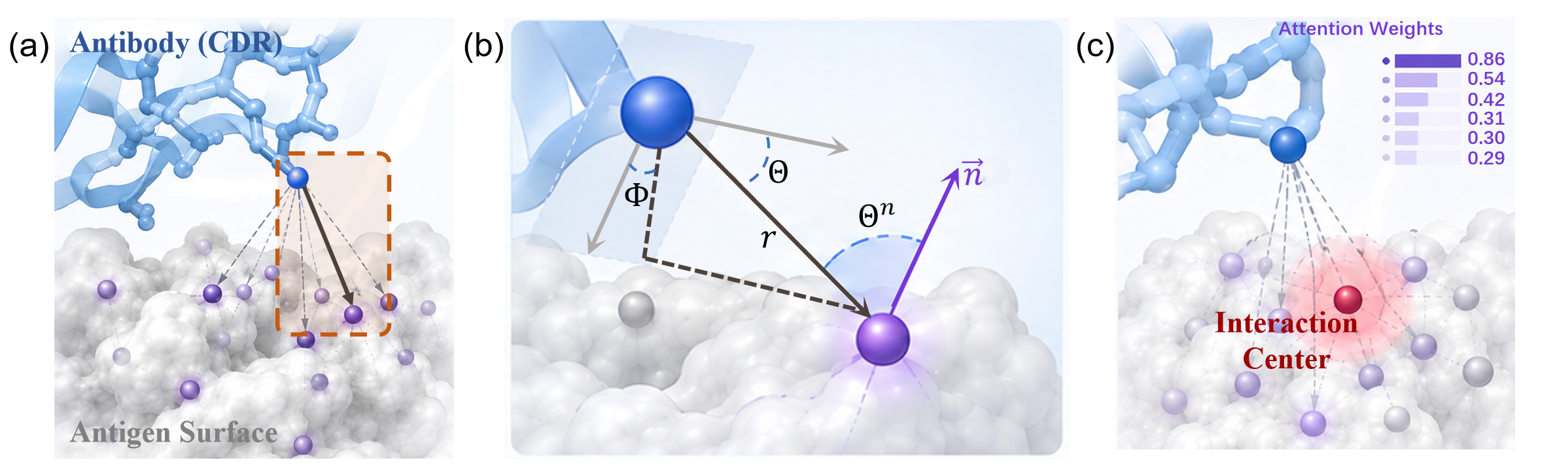}
    \caption{\textbf{Overview of local interface representation and adaptive attention.}
    \textbf{(a)} Global interaction view between the antibody (CDR) residue and the antigen surface mesh.
    \textbf{(b)} Orientation-aware geometric encoding expressed in the antibody-residue local frame, capturing distance ($r$), spatial direction ($\Theta, \Phi$), and surface normal ($\Theta^n, \vec{n}$).
    \textbf{(c)} Adaptive atom--surface attention weighting used to dynamically locate the interaction center on the antigen surface.}
    \label{fig:interface_encoding}
\end{figure*}

To capture microscopic interactions at the antibody--antigen interface, we construct an orientation-aware representation. For each antibody--antigen residue pair $e = (i, b_e)$, we relate every heavy atom of antibody residue $b_e$ to every vertex of the paired antigen surface patch $\mathcal{V}_i$. Both antibody and antigen heavy atoms are mapped into a standardized 14-slot vocabulary schema $p \in \{1,\dots,14\}$ (slots $1$--$4$ hold the backbone atoms; slots $5$--$14$ hold side-chain heavy atoms in a fixed per-type order, with unused slots padded and masked).

\textbf{Residue local frame.} For an antibody residue $a$, we construct a local coordinate frame $(\mathbf{e}_1, \mathbf{e}_2, \mathbf{e}_3)$ using its backbone heavy atoms $\mathbf{x}_{a,\mathrm{N}}$, $\mathbf{x}_{a,\mathrm{C}_\alpha}$, and $\mathbf{x}_{a,\mathrm{C}}$. Assuming non-collinear backbone atoms ($\mathbf{w} \times \mathbf{u} \neq \mathbf{0}$), we define the translation origin as $\mathbf{t}_a = \mathbf{x}_{a,\mathrm{C}_\alpha}$ and the proper rotation matrix $\mathbf{R}_a = [\mathbf{e}_1\ \mathbf{e}_2\ \mathbf{e}_3] \in \mathrm{SO}(3)$ via Gram--Schmidt orthogonalization:
\begin{equation}
\begin{aligned}
\mathbf{u} &= \mathbf{x}_{a,\mathrm{C}} - \mathbf{x}_{a,\mathrm{C}_\alpha}, 
&\quad \mathbf{e}_1 &= \frac{\mathbf{u}}{\max(\|\mathbf{u}\|_2, \epsilon)}, \\
\mathbf{w} &= \mathbf{x}_{a,\mathrm{N}} - \mathbf{x}_{a,\mathrm{C}_\alpha}, 
&\quad \mathbf{e}_2 &= \frac{\mathbf{w} - (\mathbf{e}_1^\top \mathbf{w})\mathbf{e}_1}{\max(\|\mathbf{w} - (\mathbf{e}_1^\top \mathbf{w})\mathbf{e}_1\|_2, \epsilon)}, \\
& &\quad \mathbf{e}_3 &= \mathbf{e}_1 \times \mathbf{e}_2,
\end{aligned}
\label{eq:frame}
\end{equation}
where $\epsilon = 10^{-12}$ prevents division by zero. Expressing local spatial quantities in this residue frame via $\mathbf{y}^{\mathrm{loc}} = \mathbf{R}_a^\top (\mathbf{y} - \mathbf{t}_a)$ guarantees that all downstream frame-aligned quantities, namely the geometric descriptors, attention queries, and attention weights (Eqs.~\ref{eq:geomfeat}--\ref{eq:surfattn}), are strictly $\mathrm{SE}(3)$-invariant to global rigid transformations of the molecular complex.

\textbf{Surface-aware geometric encoding.} For antibody atom slot $p$ and antigen surface vertex $j$, the relative displacement and surface normal expressed in the local frame of residue $a \equiv b_e$ are computed as $\mathbf{d}_{pj} = \mathbf{R}_a^\top (\mathbf{x}_{a,p}-\mathbf{v}_j)$ and $\vec{n}^{\mathrm{loc}}_j = \mathbf{R}_a^\top\vec{n}_j$, where $\mathbf{d}_{pj}$ points from surface vertex $j$ toward antibody atom $p$.
The descriptor combines the antibody-frame approach direction with a surface-normal incidence cosine:
\begin{equation}
\mathbf{g}_{pj} = \left[
\phi_r(r_{pj}),\;
\mathbf{A}(\Theta_{pj},\Phi_{pj}),\;
\hat{\mathbf{u}}_{pj},\;
c_{pj},\;
\mathbf{a}_{p}
\right]\in\mathbb{R}^{40},
\label{eq:geomfeat}
\end{equation}
where $r_{pj}=\|\mathbf{d}_{pj}\|_2$, $\hat{\mathbf{u}}_{pj}=\mathbf{d}_{pj}/\max(r_{pj},\epsilon)$, and $(r_{pj},\Theta_{pj},\Phi_{pj})$ are the spherical coordinates of $\mathbf{d}_{pj}$; $\phi_r:\mathbb{R}\rightarrow\mathbb{R}^{16}$ is a Gaussian radial basis~\citep{schutt2017schnet}, $\mathbf{A}(\Theta,\Phi)=[\sin\Theta,\cos\Theta,\sin\Phi,\cos\Phi]^{\top}$ a periodic directional encoding, smooth away from the polar axis of the residue frame ($\Theta=0,\pi$, where $\Phi$ is undefined), $\mathbf{a}_{p}\in\mathbb{R}^{16}$ the atom-type embedding, and $c_{pj}=\hat{\mathbf{u}}_{pj}^{\top}\vec{n}^{\mathrm{loc}}_j=\cos\Theta^{n}_{pj}$ the frame-independent approach--normal cosine, with $\Theta^{n}_{pj}$ the angle between the approach direction and the surface normal.
The direction channels expose what scalar distances cannot: two contacts with $\mathbf{d}_1 = (r,0,0)^\top$ and $\mathbf{d}_2 = (0,r,0)^\top$ have equal distance but different orientation in the antibody frame, and are separated by $\mathbf{A}(\Theta,\Phi)$ and $\hat{\mathbf{u}}_{pj}$; and while a single $c_{pj}$ does not encode the surface-normal azimuth, aggregating $c_{pj}$ across the contacts of the patch provides incidence information about the local surface orientation. All atom--surface quantities are \emph{edge-local}: computed independently for each pairing $(i, b_e)$ in the frame of $a$, never pooled across antibody residues or local frames, and mixed across edges only through the invariant edge messages $\mathbf{m}_e$ at the node level.

\subsection{Adaptive Atom--Surface Interaction and Gated Coordinate Update}
\label{sec:interaction}

We aggregate fine-grained atom--surface interactions using a two-level attention scheme.

\textbf{Atom-level aggregation.} Atom-wise interaction features are pooled into surface vertices via invariant attention weighting~\citep{fuchs2020se3transformer}:
\begin{equation}
\bar{\mathbf{g}}^{(e)}_j = \sum_{p\in a} \alpha_{pj}\mathbf{g}_{pj},
\qquad
\alpha_{pj} = \mathrm{softmax}_{p} \left(f(\mathbf{g}_{pj})\right),
\label{eq:atompool}
\end{equation}
where $f$ is a multi-layer perceptron.

\textbf{Surface-level attention and interface edge readout.} Content queries~\citep{vaswani2017attention} are constructed from the frame-aligned local atom positions $\mathbf{q}_p^{\mathrm{loc}} = \mathbf{R}_a^\top(\mathbf{x}_{a,p}-\mathbf{t}_a) \in \mathbb{R}^3$:
\begin{equation}
\beta^{(e)}_{pj} = \mathrm{softmax}_{j} \left( \frac{(\mathbf{W}_q\mathbf{q}_p^{\mathrm{loc}})^\top \mathbf{W}_k\bar{\mathbf{g}}^{(e)}_j}{\sqrt d} \right).
\label{eq:surfattn}
\end{equation}
Here, both the query $\mathbf{q}_p^{\mathrm{loc}}$ and the keys $\bar{\mathbf{g}}^{(e)}_j$ are defined within the residue-local frame of $a$.

Each atom slot retrieves a surface feature vector $\mathbf{f}_p = \sum_j \beta^{(e)}_{pj} \mathbf{W}_v \bar{\mathbf{g}}^{(e)}_j$, which is averaged to form a residue-level interaction summary $\bar{\mathbf{f}} = \frac{1}{|a|}\sum_{p\in a}\mathbf{f}_p$. Its normalized embedding $\mathbf{r}_e = \mathbf{W}_r \left( \bar{\mathbf{f}} / (\|\bar{\mathbf{f}}\|_2 + \epsilon) \right)$ is integrated into the interface edge message:
\begin{equation}
\mathbf{m}_e = \mathrm{MLP}_e \left( [\mathbf{h}_i \parallel \mathbf{h}_{b_e} \parallel \mathbf{r}_e] \right).
\label{eq:edgemessage}
\end{equation}
The invariant message $\mathbf{m}_e$ updates residue node states $\mathbf{h}_i$ and parameterizes scalar gating functions for spatial updates~\citep{satorras2021egnn}.

\textbf{Attended surface centroid and transient coordinate update.} Averaging $\beta^{(e)}_{pj}$ over the valid (unmasked) atoms of residue $a$ yields the marginal vertex distribution $\beta^{(e)}_j = \frac{1}{|a|}\sum_{p\in a}\beta^{(e)}_{pj}$ and the attended surface centroid $\bar{\mathbf{v}}_e = \sum_j \beta_j^{(e)} \mathbf{v}_j$.

Within the local interaction encoder, atomic positions are maintained as 14-slot coordinates: antigen atoms enter from the observed context, while all antibody coordinates are initialized from the conserved framework template and refined during decoding (full state provenance in Appendix~\ref{app:recurrence}). For an epitope residue $i$, its transient atomic coordinates $\tilde{\mathbf{x}}_{i,p} \in \mathbb{R}^3$ are updated across encoder layers via a gated message-passing scheme:
\begin{equation}
\tilde{\mathbf{x}}_{i,p} \leftarrow \tilde{\mathbf{x}}_{i,p} + \frac{1}{|\mathcal{N}_i|} \sum_{e\in\mathcal{N}_i} \psi_p(\mathbf{m}_e) \bigl(\mathbf{x}_{a,p}-\bar{\mathbf{v}}_e\bigr),
\label{eq:coordupdate}
\end{equation}
where $\mathbf{x}_{a,p}$ is the coordinate of atom slot $p$ and $\psi_p(\mathbf{m}_e)$ a slot-specific scalar gate projected from the invariant edge message $\mathbf{m}_e$. Here, $\tilde{\mathbf{x}}_{i,p}$ is a transient state that conditions downstream pairwise spatial contexts, not an explicit structure prediction. Unlike the residue-local descriptors and queries of Eqs.~(\ref{eq:geomfeat})--(\ref{eq:surfattn}), the quantities of Eq.~(\ref{eq:coordupdate}) live in the shared global frame of the current complex; the update acts on relative displacements and co-transforms with the complex under rigid motions (Appendix~\ref{supp:geometric_properties}).

\textbf{Layer-wise geometric recurrence.} The frames $(\mathbf{R}_a, \mathbf{t}_a)$, the descriptors $\mathbf{g}_{pj}$, the queries $\mathbf{q}_p^{\mathrm{loc}}$, and the attended centroids $\bar{\mathbf{v}}_e$ are all recomputed at each encoder layer from the current transient coordinates, while the mesh $\{\mathbf{v}_j\}$ stays anchored to the fixed antigen context. Each layer interleaves the surface-attention update with an equivariant inter-chain update whose pair-distance features are conditioned on the refined transient states $\tilde{\mathbf{x}}^{(\ell)}$, and the update of Eq.~(\ref{eq:coordupdate}) is $\mathrm{SE}(3)$-equivariant.

\subsection{Geometric Supervision}
\label{sec:supervision}

We use geometric objectives in the same residue-local frames as the interface
representation: FAPE aligns every atom in every residue frame and thereby
constrains global structural consistency, while the torsion terms regularize
local backbone geometry; together they act as a training-time geometric prior
that stabilizes what the interface representation learns.

\textbf{Frame-aligned point error.} Following \citet{jumper2021highly},
\begin{equation}
\mathcal{L}_{\mathrm{FAPE}} = \frac{1}{|\mathcal{F}||\mathcal{P}|}
\sum_{i\in\mathcal{F}}
\sum_{j\in\mathcal{P}}
\min
\left(
d_{\max},
\left|
\mathbf{R}_i^\top(\hat{\mathbf{x}}_j-\mathbf{t}_i)
-
\mathbf{R}_i^{\star\top}
(\mathbf{x}_j^\star-\mathbf{t}_i^\star)
\right|_2
\right).
\label{eq:fape}
\end{equation}

\textbf{Torsion and angle loss.} For backbone dihedrals $\theta \in \{\phi, \psi, \omega\}$ and bond angles $\alpha$, $\mathcal{L}_{\mathrm{torsion}}$ penalizes errors of the predicted cosines with the smooth-$\ell_1$ penalty $\mathrm{s}\ell_1$:
\begin{equation}
\mathcal{L}_{\mathrm{torsion}} = \sum_{\theta} \mathrm{s}\ell_1\!\left(\cos\hat{\theta}, \cos\theta\right) + \sum_{\alpha} \mathrm{s}\ell_1\!\left(\cos\hat{\alpha}, \cos\alpha\right).
\label{eq:torsion}
\end{equation}

\textbf{Total objective.} $\mathcal{L}_{\mathrm{seq}}$ is the cross-entropy of
masked-token predictions summed over decoding rounds. The structure term
comprises a Kabsch-aligned~\citep{kabsch1976rotation} full-atom coordinate loss
$\mathcal{L}_{x}$, smooth-$\ell_1$ bond-length losses $\mathcal{L}_{\mathrm{bond}}$ on
backbone and sidechain bonds, $\mathcal{L}_{\mathrm{FAPE}}$, and $\mathcal{L}_{\mathrm{torsion}}$. The docking term supervises the interface
coordinates of the shadow paratope ($\mathcal{L}_{\mathrm{SP}}$) and the
predicted inter-chain edge-distance map ($\mathcal{L}_{\mathrm{ed}}$). These
terms follow the implementation of \citet{kong2023end}. An auxiliary
smooth-$\ell_1$ term $\mathcal{L}_{\mathrm{pRMSD}}$ additionally trains a
per-residue RMSD prediction head on the antibody residues. The total
objective is
\begin{equation}
\begin{aligned}
\mathcal{L} = \lambda_{\mathrm{seq}} \mathcal{L}_{\mathrm{seq}}
& + \underbrace{\lambda_{x} \mathcal{L}_{x} + \lambda_{\mathrm{bond}} \mathcal{L}_{\mathrm{bond}} + \lambda_{\mathrm{FAPE}} \mathcal{L}_{\mathrm{FAPE}} + \lambda_{\mathrm{torsion}} \mathcal{L}_{\mathrm{torsion}}}_{\mathrm{structure}} \\
& + \underbrace{\lambda_{\mathrm{SP}} \mathcal{L}_{\mathrm{SP}} + \lambda_{\mathrm{ed}} \mathcal{L}_{\mathrm{ed}}}_{\mathrm{docking}}
+ \underbrace{\lambda_{\mathrm{pRMSD}} \mathcal{L}_{\mathrm{pRMSD}}}_{\mathrm{auxiliary}},
\end{aligned}
\label{eq:total_loss}
\end{equation}

\subsection{Generation across Modeling Tasks}
\label{sec:generation}

AbGaze decodes the sequence by stochastically revealing masked
positions~\citep{ghazvininejad2019maskpredict} under a linear unmasking
schedule, refining the coordinates conditioned on the partially
revealed sequence: the target region starts fully masked with antibody
coordinates from the conserved framework template, and each of the $K$
reveal rounds refines the structure, predicts every still-masked
position, and commits each independently,
\begin{equation}
p_\theta\!\left(x^{(s)}_i \,\middle|\, x^{(s-1)}, c\right)
= \left(1-\rho_s\right)\delta_{\mathrm{m}}
+ \rho_s\,\mathrm{softmax}\!\left(\mathbf{z}^{(s-1)}_i/\tau\right),
\qquad
\rho_s=\frac{1}{K-s+1},\quad \rho_K=1,
\label{eq:reverse}
\end{equation}
where $\mathrm{m}$ is the mask token, $x^{(s)}$ the decoding state
after reveal round $s$, $c$ the antigen and committed context,
$\mathbf{z}^{(s-1)}_i$ the round-$s$ decoder logits, and $K$ the number
of reveal rounds. Revealed tokens are never re-masked, the final round
commits the rest, and the temperature $\tau$ controls candidate
diversity. Training uses the masked-prediction objective, a member of
the simplified masked-cross-entropy family used to train
absorbing-state diffusion \citep{sahoo2024masked}, with the corruption
level set by an annealed context curriculum rather than a uniform
prior; Appendix~\ref{app:generation} states its precise scope.

Masking any subset of CDRs configures joint sequence--structure design;
observing the full antibody sequence yields complex structure
prediction from template initialization; and affinity optimization
tunes the template-initialization noise through the frozen generator,
guided by a pretrained $\Delta\Delta G$ regressor as a differentiable
proxy with a KL trust region toward $\mathcal{N}(0,\mathbf{I})$.
Across all four tasks, the same interface
representation, attention, and equivariant refinement apply: within the
shared generator there is no task-specific component or loss term, and
the tasks differ only in masking, initialization, and sampling or
optimization configurations; the sole component fitted outside the
generator is the frozen $\Delta\Delta G$ regressor, which
contributes no term to the generator's objective (Appendix~\ref{app:impl}).

\section{Experiments}
We evaluate AbGaze on the four settings served by the shared interface
representation: all-CDR design, CDR-H3 design, complex structure
prediction, and affinity optimization.

\subsection{Setup}
\label{sec:setup}
\textbf{Data and benchmark.}
Training and evaluation follow the data pipeline of \citet{kong2023end}
(details in Appendix~\ref{app:data}): SAbDab~\citep{dunbar2014sabdab}
complexes clustered by CDR sequence identity~\citep{kong2022conditional}.
To rule out cross-task leakage, we hold out the single benchmark RAbD
\citep{adolf2018rabd} (60 complexes), remove every overlapping cluster
from training, and evaluate \emph{all four} tasks exclusively on it. No
published numbers are copied; all baselines are re-run under this
protocol, with the structure-prediction and affinity baselines locally
re-evaluated (Appendix~\ref{app:protocol}).

\textbf{Baselines.}
We compare against RosettaAb~\citep{adolf2018rabd},
DiffAb~\citep{luo2022antigen}, MEAN~\citep{kong2022conditional},
HERN~\citep{jin2022antibody}, dyMEAN~\citep{kong2023end}, and
AbFlow~\citep{wang2026abflow}. Models that fill CDRs on a docked backbone
are standardized into the four-stage IgFold~\citep{ruffolo2023igfold},
HDock~\citep{yan2020hdock}, generation, and Rosetta pipeline of
\citet{kong2023end,wang2026abflow}.

\textbf{Metrics.}
Sequence quality is measured by AAR and its contact-restricted variant
CAAR; structure quality by TM-score~\citep{zhang2004tmscore},
lDDT~\citep{mariani2013lddt}, and C$_\alpha$ RMSD after alignment;
interface quality by DockQ~\citep{basu2016dockq}; and affinity by the
best $\Delta\Delta G$ (via a shared pretrained regressor), the
improvement percentage (IMP, fraction of candidates with
$\Delta\Delta G < 0$), and the mutation count $\Delta L$. Evaluation
must specify how the distribution over designs is consumed; the
protocol is fixed per task, with every method sampled and selected
identically within the design and affinity comparisons. \emph{Design}
(all-CDR and CDR-H3): five samples per target at $\tau{=}0.5$,
reporting the candidate with the highest AAR against the native, an
oracle selection applied identically to every method. \emph{Complex
structure prediction}: ten stochastic draws (random shadow-paratope
initialization), keeping the one with the lowest model-predicted
per-residue RMSD; no ground truth is involved. \emph{Affinity
optimization}: thirty optimized candidates per target, reporting the
best $\Delta\Delta G$; IMP is computed over all candidates. The budget
contributes little: best-of-five adds only 2.2 AAR points over a
single deterministic design (Appendix~\ref{app:gains_without_sampling}).

\subsection{All-CDR Design}
\label{sec:design_all}
Table~\ref{tab:main_results} evaluates the joint design of all six CDRs.
The \emph{All} AAR is pooled over the residues of the six CDRs, while the
\emph{All} RMSD is the C$_\alpha$ RMSD of the entire antibody. The pooled
AAR reaches 66.2\%, a 10\% relative gain over the strongest baseline.
Relative to
the strongest baseline on each CDR, AbGaze improves AAR by 7.1\% and
reduces RMSD by 14.9\% on average over the six loops, with the largest
gains on the conformationally variable L3 and H3: L3 AAR reaches 0.69
($+19\%$) and the H3 C$_\alpha$ RMSD falls to 1.65\,\AA{} ($-10\%$).
Interface quality follows, with a DockQ of 0.422 vs.\ 0.396 ($+6.6\%$).

\begin{table}[h]
\centering
\small
\setlength{\tabcolsep}{10pt}
\caption{Performance comparison on all-CDR antibody design.
$\uparrow$ indicates higher is better, while $\downarrow$ indicates
lower is better. Bold denotes the best performance.}
\label{tab:main_results}
\begin{tabular}{@{}cc|ccc@{}}
\toprule
\textbf{Metric} & \textbf{Item}
& \textbf{AbGaze}
& \textbf{AbFlow}
& \textbf{dyMEAN} \\
\midrule
\multirow{7}{*}{\textbf{AAR}$\uparrow$}
& L1 & \textbf{0.77} & 0.69 & 0.76 \\
& L2 & \textbf{0.85} & 0.82 & 0.83 \\
& L3 & \textbf{0.69} & 0.58 & 0.52 \\
& H1 & \textbf{0.79} & 0.74 & 0.76 \\
& H2 & \textbf{0.71} & 0.65 & 0.69 \\
& H3 & \textbf{0.43} & 0.38 & 0.38 \\
\addlinespace[2pt]
& All & \textbf{0.662} & 0.597 & 0.601 \\
\midrule
\multirow{7}{*}{\textbf{RMSD (CA)}$\downarrow$}
& L1 & \textbf{0.44} & 0.64 & 0.86 \\
& L2 & \textbf{0.21} & 0.25 & 0.48 \\
& L3 & \textbf{0.65} & 0.65 & 0.94 \\
& H1 & \textbf{0.52} & 0.63 & 0.63 \\
& H2 & \textbf{0.47} & 0.55 & 0.71 \\
& H3 & \textbf{1.65} & 1.83 & 2.45 \\
\addlinespace[2pt]
& All & \textbf{1.052} & 1.104 & 1.357 \\
\midrule
\multirow{3}{*}{\textbf{Structure}}
& DockQ$\uparrow$ & \textbf{0.422} & 0.379 & 0.396 \\
& LDDT$\uparrow$ & \textbf{0.831} & 0.815 & 0.803 \\
& TM-Score$\uparrow$ & \textbf{0.973} & 0.971 & 0.965 \\
\bottomrule
\end{tabular}
\end{table}

The concurrent improvements across heavy- and light-chain CDRs suggest
that explicit orientation and distance cues help coordinate multiple
flexible loops: resolving each residue's contacts in its own local
frame against the epitope surface supports joint optimization of
paratope sequence and backbone geometry, keeping adjacent loops
coherent while aligning the paratope with the antigen.

\subsection{CDR-H3 Design}
\label{sec:design_h3}
Table~\ref{tab:cdrh3_design} evaluates CDR-H3 design on RAbD, the most
widely optimized loop given H3's decisive role in antigen recognition. On
global backbone metrics, most end-to-end baselines reach comparably high
performance ($\ge 0.84$ lDDT, $\ge 0.97$ TM-score), indicating that coarse
loop topology is largely well-captured. AbGaze leads across sequence,
interface, and structural metrics, reaching an AAR of 45.60\% ($+4.5\%$),
a contact-restricted recovery of 32.20\% ($+11.8\%$), a DockQ of
0.443 ($+3.5\%$), and the
lowest C$_\alpha$ RMSD (8.06), all
relative to the strongest baseline. The gains concentrate at the
interface itself, suggesting that encoding the distance, direction, and angle to the
surface normal at which each atom approaches the epitope lets the model
align local paratope geometry with epitope constraints.

\begin{table}[h]
\centering
\caption{Performance comparison on CDR-H3 antibody design on the RAbD benchmark.}
\label{tab:cdrh3_design}
\small
\begin{tabularx}{\linewidth}{@{}>{\centering\arraybackslash}X*{6}{>{\centering\arraybackslash}X}@{}}
\toprule
\textbf{Method}
& \textbf{AAR} $\uparrow$
& \textbf{TMscore} $\uparrow$
& \textbf{lDDT} $\uparrow$
& \textbf{CAAR} $\uparrow$
& \textbf{RMSD} $\downarrow$
& \textbf{DockQ} $\uparrow$ \\
\midrule
RosettaAb & 32.31\% & 0.9717 & 0.8272 & 14.58\% & 17.70 & 0.137 \\
DiffAb    & 35.31\% & 0.9695 & 0.8281 & 22.17\% & 23.24 & 0.158 \\
MEAN      & 37.38\% & 0.9688 & 0.8252 & 24.11\% & 17.30 & 0.162 \\
HERN      & 32.65\% & --     & --     & 19.27\% & 9.15  & 0.294 \\
dyMEAN    & 43.65\% & 0.9726 & 0.8454 & 28.11\% & 8.11  & 0.409 \\
AbFlow    & 42.10\% & \textbf{0.9735} & \textbf{0.8518} & 28.80\% & 8.45 & 0.428 \\
\midrule
\textbf{AbGaze}
& \textbf{45.60\%}
& 0.9730
& 0.8455
& \textbf{32.20\%}
& \textbf{8.06}
& \textbf{0.443} \\
\bottomrule
\end{tabularx}
\end{table}

\subsection{Complex Structure Prediction}
\label{sec:predict}
Table~\ref{tab:struct_prediction} evaluates prediction from sequences
alone against the docking pipeline HDock, the hierarchical refinement of
HERN, and the end-to-end baselines. Global antibody structure is
near-saturated for all end-to-end methods (TM-score $\geq 0.97$); the
differences appear precisely at the interface, where AbGaze attains the
best DockQ (0.435) and RMSD (8.03). The simultaneous DockQ and RMSD gains
indicate that the representation better resolves the paratope's spatial
arrangement relative to the antigen surface: the distance-, direction-,
and orientation-structure of the static binding mode on this benchmark.
These gains show that the representation characterizes the binding mode
accurately.

\begin{table}[h]
\centering
\small
\caption{Performance comparison on antigen--antibody complex structure prediction.}
\label{tab:struct_prediction}
\begin{tabularx}{\linewidth}{
    @{}
    >{\centering\arraybackslash}X
    *{4}{>{\centering\arraybackslash}X}
    @{}
}
\toprule
\textbf{Model}
& \textbf{TMscore} $\uparrow$
& \textbf{lDDT} $\uparrow$
& \textbf{RMSD} $\downarrow$
& \textbf{DockQ} $\uparrow$ \\
\midrule
HDock  & 0.9723 & 0.8503 & 18.46 & 0.170 \\
HERN   & 0.9722 & 0.8441 & 10.19 & \underline{0.424} \\
dyMEAN & \underline{0.9730} & \underline{0.8568} & 9.04 & 0.409 \\
AbFlow & 0.9720 & 0.8526 & \underline{8.66} & 0.419 \\

\midrule
\textbf{AbGaze} & \textbf{0.9735} & \textbf{0.8572} & \textbf{8.03} & \textbf{0.435} \\
\bottomrule
\end{tabularx}
\end{table}

\subsection{Affinity Optimization}
\label{sec:affinity}
Table~\ref{tab:affinity} compares affinity optimization on RAbD with the
same frozen generator and the same $\Delta\Delta G$ regressor for all
methods. AbGaze reaches the best $\Delta\Delta G$ ($-11.10$) and by far
the highest improvement rate (IMP 70.6\%, $+17.3$ points over dyMEAN and
$+17.8$ over AbFlow): almost three quarters of the optimized candidates
receive a negative predicted $\Delta\Delta G$ from the shared scorer. The mutations it requests are more numerous ($\Delta L$
8.07 vs.\ 4.25 for dyMEAN); we regard this as an acceptable trade, since
IMP already scores success per candidate. That gradient ascent through
the shared representation alone suffices to steer affinity supports the
central claim of \S\ref{sec:representation}: the orientation-aware
encoding captures the interfacial context---proximity, relative
orientation, and the contribution of each contact under its surrounding
geometry---well enough that optimizing toward this representation, rather
than merely fitting the design objective, transfers to a downstream
affinity criterion. Together with the binding-mode results in
prediction, this is consistent with the representation characterizing
how the antibody binds, and how strongly, more faithfully than
distance- or surface-based encodings.

\begin{table}[h]
\centering
\caption{Comparison of affinity optimization performance.}
\label{tab:affinity}
\small
\begin{tabularx}{\linewidth}{
    @{}
    >{\centering\arraybackslash}X
    *{3}{>{\centering\arraybackslash}X}
    @{}
}
\toprule
\textbf{Method}
& \textbf{Best $\Delta\Delta G$} $\downarrow$
& \textbf{IMP (\%)} $\uparrow$
& \textbf{$\Delta L$} $\downarrow$ \\
\midrule
DiffAb & -3.29 & 38.8 & 5.62 \\
dyMEAN & -4.47  & 53.3 & \textbf{4.25} \\
AbFlow & -9.31  & 52.8 & 6.98 \\

\midrule
\textbf{AbGaze} & \textbf{-11.10} & \textbf{70.6} & 8.07 \\
\bottomrule
\end{tabularx}
\end{table}

\section{Ablation}
We ablate the orientation-aware interaction module (Variant~A) and the residue-local geometric loss (Variant~B) on RAbD all-CDR design.

\textbf{Orientation-aware interaction.} Variant~(A) replaces the local interface representation (\S\ref{sec:representation}--\ref{sec:interaction}) with standard distance-based message passing. Removing explicit spatial and normal directionality primarily weakens interface packing (DockQ \(-0.040\), antibody C\(\alpha\) RMSD \(+0.061\)~\AA) and reduces overall sequence recovery (AAR \(-4.1\%\)), showing that atom--surface orientation carries information beyond scalar distances.

\textbf{Local-frame geometric supervision.} Variant~(B) omits the residue-local FAPE and dihedral terms (\S\ref{sec:supervision}), relying solely on global coordinate and auxiliary losses. While interface docking is essentially unchanged (DockQ \(+0.012\)), backbone structural quality degrades significantly (antibody C\(\alpha\) RMSD \(+0.147\)~\AA, lDDT \(-0.043\), TM-Score \(-0.009\)). The local-frame terms thus act as global geometric regularizers that constrain the relative orientation of adjacent residues and stabilize the overall fold.

\begin{table}[h]
    \centering
    \setlength{\tabcolsep}{4pt}
    \caption{Ablation study on all-CDR antibody design.}
    \label{tab:ablation}
    \small
    \begin{tabular}{l *{5}{c}}
        \toprule
        \textbf{Variant}
        & \textbf{AAR} \(\uparrow\)
        & \textbf{RMSD} \(\downarrow\)
        & \textbf{DockQ} \(\uparrow\)
        & \textbf{lDDT} \(\uparrow\)
        & \textbf{TMscore} \(\uparrow\) \\
        \midrule
        \textbf{AbGaze (Full)}
        & \textbf{66.2\%} & \textbf{1.052} & \textbf{0.422} & \textbf{0.831} & \textbf{0.973} \\
        \midrule
        (A) w/o Orient.-Aware Attn.
        & \(-4.1\%\) & \(+0.061\) & \(-0.040\) & \(-0.007\) & \(-0.003\) \\
        \addlinespace[2pt]
        (B) w/o Local-Frame Supv.
        & \(-1.2\%\) & \(+0.147\) & \(+0.012\) & \(-0.043\) & \(-0.009\) \\
        \bottomrule
    \end{tabular}
\end{table}

\section{Conclusion} 
We presented AbGaze, a unified framework for antigen--antibody interface
modeling built on an orientation-aware local representation and adaptive
atom--surface attention. By anchoring distance, direction, and
surface-normal geometry to residue-local frames, AbGaze explicitly
encodes how each antibody atom approaches the antigen surface, and its
two-level attention learns which local interactions matter under each
interfacial context. A single
model covers all four tasks: relative to the second-best method, it
improves amino-acid recovery by 7.1\% and reduces per-CDR $C_\alpha$
RMSD by 14.9\% on average over the six CDRs, improves DockQ by 6.6\%,
and raises the affinity improvement rate (IMP) by 32.5\%. These results
support the premise that explicitly characterizing the relative spatial
relationships between paratope and epitope in antibody-residue local
frames outperforms leaving them implicit in scalar distances or
surface-intrinsic features. Next steps include extending the
representation to flexible antigens, validating designs experimentally,
and incorporating developability into the optimization objective.

\section*{Acknowledgements}
This research is supported by the Innovative Drug Research and Development National Science and Technology Major Project (No.2025ZD1803101),  the Wuxi Research Institute of Applied Technologies, Tsinghua University (Grant 20242001120), and PharMolix Inc.

\bibliographystyle{iclr2027_conference}
\bibliography{iclr2027_conference}

@article{chan2025fifty,
  title={Fifty years of monoclonals: the past, present and future of antibody therapeutics},
  author={Chan, Andrew C. and Martyn, Greg D. and Carter, Paul J.},
  journal={Nature Reviews Immunology},
  volume={25},
  pages={745--765},
  year={2025},
  doi={10.1038/s41577-025-01207-9}
}

@article{selaculang2013recognition,
  title={The structural basis of antibody-antigen recognition},
  author={Sela-Culang, Inbal and Kunik, Vered and Ofran, Yanay},
  journal={Frontiers in Immunology},
  volume={4},
  pages={302},
  year={2013},
  doi={10.3389/fimmu.2013.00302}
}

@article{weitzner2015h3,
  title={The origin of CDR H3 structural diversity},
  author={Weitzner, Brian D. and Dunbrack Jr., Roland L. and Gray, Jeffrey J.},
  journal={Structure},
  volume={23},
  number={2},
  pages={302--311},
  year={2015},
  doi={10.1016/j.str.2014.11.010}
}

@article{jin2022antibody,
  title={Antibody-Antigen Docking and Design via Hierarchical Equivariant Refinement},
  author={Jin, Wengong and Barzilay, Regina and Jaakkola, Tommi S.},
  journal={arXiv preprint arXiv:2207.06616},
  year={2022}
}

@inproceedings{luo2022antigen,
  title={Antigen-Specific Antibody Design and Optimization with Diffusion-Based Generative Models for Protein Structures},
  author={Luo, Shitong and Yang, Kevin K. and Xu, Minkai and Wu, Zhuoran and Xie, Pengtao and Jin, Wengong and Tang, Bowen and Peng, Jian and Ma, Jianzhu},
  booktitle={Advances in Neural Information Processing Systems},
  volume={35},
  pages={9754--9767},
  year={2022}
}

@inproceedings{kong2023end,
  title={End-to-End Full-Atom Antibody Design},
  author={Kong, Xiangzhe and Huang, Wenbing and Liu, Yang},
  booktitle={Proceedings of the 40th International Conference on Machine Learning},
  pages={17409--17429},
  year={2023},
  volume={202},
  series={Proceedings of Machine Learning Research},
  publisher={PMLR}
}

@inproceedings{wang2025iggm,
  title={IgGM: A Generative Model for Functional Antibody and Nanobody Design},
  author={Wang, Rubo and Wu, Fandi and Gao, Xingyu and Wu, Jiaxiang and Zhao, Peilin and Yao, Jianhua},
  booktitle={The Thirteenth International Conference on Learning Representations},
  year={2025},
  url={https://openreview.net/forum?id=zmmfsJpYcq}
}

@inproceedings{tan2025dyab,
  title={dyAb: Flow Matching for Flexible Antibody Design with AlphaFold-driven Pre-binding Antigen},
  author={Tan, Cheng and Zhang, Yijie and Gao, Zhangyang and Huang, Yufei and Lin, Haitao and Wu, Lirong and Wu, Fandi and Blanchette, Mathieu and Li, Stan Z.},
  booktitle={Proceedings of the AAAI Conference on Artificial Intelligence},
  volume={39},
  pages={782--790},
  year={2025},
  doi={10.1609/aaai.v39i1.32061}
}

@inproceedings{wang2026abflow,
  title={AbFlow: End-to-end Paratope-Centric Antibody Design by Interaction Enhanced Flow Matching},
  author={Wang, Wenda and Zhang, Yang and Wei, Zhewei and Huang, Wenbing},
  booktitle={Proceedings of the 32nd ACM SIGKDD Conference on Knowledge Discovery and Data Mining},
  year={2026},
  doi={10.1145/3770854.3780296}
}

@article{gainza2020deciphering,
  title={Deciphering interaction fingerprints from protein molecular surfaces using geometric deep learning},
  author={Gainza, Pablo and Sverrisson, Freyr and Monti, Federico and Rodol{\`a}, Emanuele and Boscaini, Davide and Bronstein, Michael M. and Correia, Bruno E.},
  journal={Nature Methods},
  volume={17},
  pages={184--192},
  year={2020},
  doi={10.1038/s41592-019-0666-6}
}

@article{gainza2023masifseed,
  title={De novo design of protein interactions with learned surface fingerprints},
  author={Gainza, Pablo and Wehrle, Sarah and Van Hall-Beauvais, Alexandra and Marchand, Anthony and Scheck, Andreas and Harteveld, Zander and Buckley, Stephen and Ni, Dongchun and Tan, Shuguang and Sverrisson, Freyr and Goverde, Casper and Turelli, Priscilla and Raclot, Charl{\`e}ne and Teslenko, Alexandra and Pacesa, Martin and Rosset, St{\'e}phane and Georgeon, Sandrine and Marsden, Jane and Petruzzella, Aaron and Liu, Kefang and Xu, Zepeng and Chai, Yan and Han, Pu and Gao, George F. and Oricchio, Elisa and Fierz, Beat and Trono, Didier and Stahlberg, Henning and Bronstein, Michael and Correia, Bruno E.},
  journal={Nature},
  volume={617},
  pages={176--184},
  year={2023},
  doi={10.1038/s41586-023-05993-x}
}

@article{kong2022conditional,
  title={Conditional Antibody Design as 3D Equivariant Graph Translation},
  author={Kong, Xiangzhe and Huang, Wenbing and Liu, Yang},
  journal={arXiv preprint arXiv:2208.06073},
  year={2022}
}

@article{dai2021pinet,
  title={Protein interaction interface region prediction by geometric deep learning},
  author={Dai, Bowen and Bailey-Kellogg, Chris},
  journal={Bioinformatics},
  volume={37},
  number={17},
  pages={2580--2588},
  year={2021},
  doi={10.1093/bioinformatics/btab154}
}

@article{krapp2023pesto,
  title={PeSTo: parameter-free geometric deep learning for accurate prediction of protein binding interfaces},
  author={Krapp, Lucien F. and Abriata, Luciano A. and Cort{\'e}s Rodriguez, Fabio and Dal Peraro, Matteo},
  journal={Nature Communications},
  volume={14},
  number={1},
  pages={2175},
  year={2023},
  doi={10.1038/s41467-023-37701-8}
}

@article{chen2024probid,
  title={ProBID-Net: a deep learning model for protein-protein binding interface design},
  author={Chen, Zhihang and Ji, Menglin and Qian, Jie and Zhang, Zhe and Zhang, Xiangying and Gao, Haotian and Wang, Haojie and Wang, Renxiao and Qi, Yifei},
  journal={Chemical Science},
  volume={15},
  number={47},
  pages={19977--19990},
  year={2024},
  doi={10.1039/D4SC02233E}
}

@article{lawrence1993shape,
  author  = {Lawrence, Michael C. and Colman, Peter M.},
  title   = {Shape Complementarity at Protein/Protein Interfaces},
  journal = {Journal of Molecular Biology},
  year    = {1993},
  volume  = {234},
  number  = {4},
  pages   = {946--950},
  doi     = {10.1006/jmbi.1993.1648}
}

@article{mccoy1997electrostatic,
  author  = {McCoy, A. J. and Epa, V. Chandana and Colman, P. M.},
  title   = {Electrostatic Complementarity at Protein/Protein Interfaces},
  journal = {Journal of Molecular Biology},
  year    = {1997},
  volume  = {268},
  number  = {2},
  pages   = {570--584},
  doi     = {10.1006/jmbi.1997.0987}
}

@article{yang2003cooperative,
  author  = {Yang, Jianying and Swaminathan, Chittoor P. and Huang, Yuping and Guan, Rongjin and Cho, Sangwoo and Kieke, Michele C. and Kranz, David M. and Mariuzza, Roy A. and Sundberg, Eric J.},
  title   = {Dissecting Cooperative and Additive Binding Energetics in the Affinity Maturation Pathway of a Protein-Protein Interface},
  journal = {Journal of Biological Chemistry},
  year    = {2003},
  volume  = {278},
  number  = {50},
  pages   = {50412--50421},
  doi     = {10.1074/jbc.M306848200}
}

@article{jumper2021highly,
  author  = {Jumper, John and Evans, Richard and Pritzel, Alexander and Green, Tim and Figurnov, Michael and Ronneberger, Olaf and Tunyasuvunakool, Kathryn and Bates, Russ and Z{\'i}dek, Augustin and Potapenko, Anna and Bridgland, Alex and Meyer, Clemens and Kohl, Simon A. A. and Ballard, Andrew J. and Cowie, Andrew and Romera-Paredes, Bernardino and Nikolov, Stanislav and Jain, Rishub and Adler, Jonas and Back, Trevor and Petersen, Stig and Reiman, David and Clancy, Ellen and Zielinski, Michal and Steinegger, Martin and Pacholska, Michalina and Berghammer, Tamas and Bodenstein, Sebastian and Silver, David and Vinyals, Oriol and Senior, Andrew W. and Kavukcuoglu, Koray and Kohli, Pushmeet and Hassabis, Demis},
  title   = {Highly Accurate Protein Structure Prediction with AlphaFold},
  journal = {Nature},
  year    = {2021},
  volume  = {596},
  number  = {7873},
  pages   = {583--589},
  doi     = {10.1038/s41586-021-03819-2}
}

@article{geogad2026,
  author  = {Wei, Songjian and Zhang, Jinxiong and Chen, Yan and Tang, Chunyan and Tan, Jiayang},
  title   = {GeoGAD: geometry-aware antibody design framework for complementarity-determining region precision engineering},
  journal = {Bioinformatics},
  year    = {2026},
  volume  = {42},
  number  = {2},
  pages   = {btag042},
  doi     = {10.1093/bioinformatics/btag042}
}

@article{sanner1996reduced,
  author  = {Sanner, Michel F. and Olson, Arthur J. and Spehner, Jean-Claude},
  title   = {Reduced surface: An efficient way to compute molecular surfaces},
  journal = {Biopolymers},
  year    = {1996},
  volume  = {38},
  number  = {3},
  pages   = {305--320},
  doi     = {10.1002/(SICI)1097-0282(199603)38:3<305::AID-BIP4>3.0.CO;2-Y}
}

@article{adolf2018rabd,
  author  = {Adolf-Bryfogle, Jared and Kalyuzhniy, Oleks and Kubitz, Michael and Weitzner, Brian D. and Hu, Xiaozhen and Adachi, Yumiko and Schief, William R. and Dunbrack, Roland L.},
  title   = {RosettaAntibodyDesign (RAbD): A general framework for computational antibody design},
  journal = {PLOS Computational Biology},
  year    = {2018},
  volume  = {14},
  number  = {4},
  pages   = {e1006112},
  doi     = {10.1371/journal.pcbi.1006112}
}

@article{basu2016dockq,
  author  = {Basu, Sankar and Wallner, Bj{\"o}rn},
  title   = {DockQ: A Quality Measure for Protein-Protein Docking Models},
  journal = {PLOS ONE},
  year    = {2016},
  volume  = {11},
  number  = {8},
  pages   = {e0161879},
  doi     = {10.1371/journal.pone.0161879}
}

@inproceedings{austin2021structured,
  author    = {Austin, Jacob and Johnson, Daniel D. and Ho, Jonathan and Tarlow, Daniel and van den Berg, Rianne},
  title     = {Structured Denoising Diffusion Models in Discrete State-Spaces},
  booktitle = {Advances in Neural Information Processing Systems},
  volume    = {34},
  pages     = {17981--17993},
  year      = {2021}
}

@inproceedings{sahoo2024masked,
  author    = {Sahoo, Subham Sekhar and Arriola, Marianne and Schiff, Yair and Gokaslan, Aaron and Marroquin, Edgar and Chiu, Justin T. and Rush, Alexander and Kuleshov, Volodymyr},
  title     = {Simple and Effective Masked Diffusion Language Models},
  booktitle = {Advances in Neural Information Processing Systems},
  volume    = {37},
  pages     = {130136--130184},
  year      = {2024},
  doi       = {10.52202/079017-4135}
}

@article{north2011cdr,
  author  = {North, Benjamin and Lehmann, Andreas and Dunbrack Jr., Roland L.},
  title   = {A new clustering of antibody {CDR} loop conformations},
  journal = {Journal of Molecular Biology},
  year    = {2011},
  volume  = {406},
  number  = {2},
  pages   = {228--256},
  doi     = {10.1016/j.jmb.2010.10.030}
}

@inproceedings{jin2022refinegnn,
  author    = {Jin, Wengong and Wohlwend, Jeremy and Barzilay, Regina and Jaakkola, Tommi},
  title     = {Iterative Refinement Graph Neural Network for Antibody Sequence-Structure Co-design},
  booktitle = {International Conference on Learning Representations},
  year      = {2022},
  url       = {https://arxiv.org/abs/2110.04624}
}

@inproceedings{martinkus2023abdiffuser,
  author    = {Martinkus, Karolis and Ludwiczak, Jan and Liang, Wei-Ching and Lafrance-Vanasse, Julien and Hotzel, Isidro and Rajpal, Arvind and Wu, Yan and Cho, Kyunghyun and Bonneau, Richard and Gligorijevic, Vladimir and Loukas, Andreas},
  title     = {{AbDiffuser}: Full-Atom Generation of In-Vitro Functioning Antibodies},
  booktitle = {Advances in Neural Information Processing Systems},
  year      = {2023},
  volume    = {36},
  url       = {https://proceedings.neurips.cc/paper_files/paper/2023/hash/801ec05b0aae9fcd2ef35c168bd538e0-Abstract-Conference.html}
}

@article{kuroda2016complementarity,
  author  = {Kuroda, Daisuke and Gray, Jeffrey J.},
  title   = {Shape complementarity and hydrogen bond preferences in protein--protein interfaces: implications for antibody modeling and protein--protein docking},
  journal = {Bioinformatics},
  year    = {2016},
  volume  = {32},
  number  = {16},
  pages   = {2451--2456},
  doi     = {10.1093/bioinformatics/btw197}
}

@inproceedings{sverrisson2021dmasif,
  author    = {Sverrisson, Freyr and Feydy, Jean and Correia, Bruno E. and Bronstein, Michael M.},
  title     = {Fast End-to-End Learning on Protein Surfaces},
  booktitle = {Proceedings of the IEEE/CVF Conference on Computer Vision and Pattern Recognition},
  year      = {2021},
  pages     = {15272--15281},
  url       = {https://openaccess.thecvf.com/content/CVPR2021/html/Sverrisson_Fast_End-to-End_Learning_on_Protein_Surfaces_CVPR_2021_paper.html}
}

@inproceedings{jing2021gvp,
  author    = {Jing, Bowen and Eismann, Stephan and Suriana, Patricia and Townshend, Raphael J. L. and Dror, Ron},
  title     = {Learning from Protein Structure with Geometric Vector Perceptrons},
  booktitle = {International Conference on Learning Representations},
  year      = {2021},
  url       = {https://arxiv.org/abs/2009.01411}
}

@inproceedings{schutt2017schnet,
  author    = {Sch{\"u}tt, Kristof T. and Kindermans, Pieter-Jan and Sauceda, Huziel E. and Chmiela, Stefan and Tkatchenko, Alexandre and M{\"u}ller, Klaus-Robert},
  title     = {{SchNet}: A continuous-filter convolutional neural network for modeling quantum interactions},
  booktitle = {Advances in Neural Information Processing Systems},
  year      = {2017},
  volume    = {30},
  url       = {https://proceedings.neurips.cc/paper/2017/hash/303ed4c69846ab36c2904d3ba8573050-Abstract.html}
}

@inproceedings{fuchs2020se3transformer,
  author    = {Fuchs, Fabian B. and Worrall, Daniel E. and Fischer, Volker and Welling, Max},
  title     = {{SE(3)}-Transformers: {3D} Roto-Translation Equivariant Attention Networks},
  booktitle = {Advances in Neural Information Processing Systems},
  year      = {2020},
  volume    = {33},
  pages     = {1970--1981},
  url       = {https://proceedings.neurips.cc/paper/2020/hash/15231a7ce4ba789d13b722cc5c955834-Abstract.html}
}

@inproceedings{vaswani2017attention,
  author    = {Vaswani, Ashish and Shazeer, Noam and Parmar, Niki and Uszkoreit, Jakob and Jones, Llion and Gomez, Aidan N. and Kaiser, {\L}ukasz and Polosukhin, Illia},
  title     = {Attention Is All You Need},
  booktitle = {Advances in Neural Information Processing Systems},
  year      = {2017},
  volume    = {30},
  pages     = {5998--6008},
  url       = {https://papers.nips.cc/paper/7181-attention-is-all-you-need}
}

@inproceedings{satorras2021egnn,
  author    = {Satorras, V{\'i}ctor Garcia and Hoogeboom, Emiel and Welling, Max},
  title     = {{E(n)} Equivariant Graph Neural Networks},
  booktitle = {Proceedings of the 38th International Conference on Machine Learning},
  series    = {Proceedings of Machine Learning Research},
  volume    = {139},
  pages     = {9323--9332},
  publisher = {PMLR},
  year      = {2021},
  url       = {https://proceedings.mlr.press/v139/satorras21a.html}
}

@article{kabsch1976rotation,
  author  = {Kabsch, Wolfgang},
  title   = {A solution for the best rotation to relate two sets of vectors},
  journal = {Acta Crystallographica Section A},
  year    = {1976},
  volume  = {32},
  number  = {5},
  pages   = {922--923},
  doi     = {10.1107/S0567739476001873}
}

@article{sircar2010snugdock,
  author  = {Sircar, Aroop and Gray, Jeffrey J.},
  title   = {{SnugDock}: Paratope Structural Optimization during Antibody-Antigen Docking Compensates for Errors in Antibody Homology Models},
  journal = {PLOS Computational Biology},
  year    = {2010},
  volume  = {6},
  number  = {1},
  pages   = {e1000644},
  doi     = {10.1371/journal.pcbi.1000644}
}

@article{ruffolo2023igfold,
  author  = {Ruffolo, Jeffrey A. and Chu, Lee-Shin and Mahajan, Sai Pooja and Gray, Jeffrey J.},
  title   = {Fast, accurate antibody structure prediction from deep learning on massive set of natural antibodies},
  journal = {Nature Communications},
  year    = {2023},
  volume  = {14},
  pages   = {2389},
  doi     = {10.1038/s41467-023-38063-x}
}

@article{yan2020hdock,
  author  = {Yan, Yumeng and Tao, Huanyu and He, Jiahua and Huang, Sheng-You},
  title   = {The {HDOCK} server for integrated protein--protein docking},
  journal = {Nature Protocols},
  year    = {2020},
  volume  = {15},
  number  = {5},
  pages   = {1829--1852},
  doi     = {10.1038/s41596-020-0312-x}
}

@article{dunbar2014sabdab,
  author  = {Dunbar, James and Krawczyk, Konrad and Leem, Jinwoo and Baker, Terry and Fuchs, Angelika and Georges, Guy and Shi, Jiye and Deane, Charlotte M.},
  title   = {{SAbDab}: the structural antibody database},
  journal = {Nucleic Acids Research},
  year    = {2014},
  volume  = {42},
  number  = {D1},
  pages   = {D1140--D1146},
  doi     = {10.1093/nar/gkt1043}
}

@article{zhang2004tmscore,
  author  = {Zhang, Yang and Skolnick, Jeffrey},
  title   = {Scoring function for automated assessment of protein structure template quality},
  journal = {Proteins: Structure, Function, and Bioinformatics},
  year    = {2004},
  volume  = {57},
  number  = {4},
  pages   = {702--710},
  doi     = {10.1002/prot.20264}
}

@article{mariani2013lddt,
  author  = {Mariani, Valerio and Biasini, Marco and Barbato, Alessandro and Schwede, Torsten},
  title   = {{lDDT}: a local superposition-free score for comparing protein structures and models using distance difference tests},
  journal = {Bioinformatics},
  year    = {2013},
  volume  = {29},
  number  = {21},
  pages   = {2722--2728},
  doi     = {10.1093/bioinformatics/btt473}
}

@article{lefranc2003imgt,
  author  = {Lefranc, Marie-Paule and Pommi{\'e}, Christelle and Ruiz, Manuel and Giudicelli, V{\'e}ronique and Foulquier, Elodie and Truong, Lisa and Thouvenin-Contet, Val{\'e}rie and Lefranc, G{\'e}rard},
  title   = {{IMGT} unique numbering for immunoglobulin and {T} cell receptor variable domains and {Ig} superfamily {V}-like domains},
  journal = {Developmental and Comparative Immunology},
  year    = {2003},
  volume  = {27},
  number  = {1},
  pages   = {55--77},
  doi     = {10.1016/S0145-305X(02)00039-3}
}

@inproceedings{ghazvininejad2019maskpredict,
  author    = {Ghazvininejad, Marjan and Levy, Omer and Liu, Yinhan and Zettlemoyer, Luke},
  title     = {{Mask-Predict}: Parallel Decoding of Conditional Masked Language Models},
  booktitle = {Proceedings of the 2019 Conference on Empirical Methods in Natural Language Processing and the 9th International Joint Conference on Natural Language Processing (EMNLP-IJCNLP)},
  year      = {2019},
  pages     = {6112--6121},
  publisher = {Association for Computational Linguistics},
  doi       = {10.18653/v1/D19-1633},
  url       = {https://aclanthology.org/D19-1633/}
}

@inproceedings{kingma2015adam,
  author    = {Kingma, Diederik P. and Ba, Jimmy Lei},
  title     = {{Adam}: A Method for Stochastic Optimization},
  booktitle = {International Conference on Learning Representations},
  year      = {2015},
  url       = {https://arxiv.org/abs/1412.6980}
}

\appendix
\appendix

\section{Extended Method Details}
\label{app:method}

This section supplements \S\ref{sec:representation}--\S\ref{sec:interaction}
with the implementation-level details behind the interface
representation: the atom-slot schema and the construction of interface
edges (Appendix~\ref{app:atomslots}), the surface-patch construction
(Appendix~\ref{app:surface}), the composition of the geometric
descriptor and the normalization axes of the two-level attention
(Appendix~\ref{app:descriptor}), and the transient coordinate states,
their layer-wise recurrence, and the $\mathrm{SE}(3)$-equivariance of
the interface update (Appendices~\ref{app:recurrence}
and~\ref{supp:geometric_properties}). Notation follows the main text.

\subsection{Atom-Slot Schema and Interface Edges}
\label{app:atomslots}

\textbf{Atom slots.}
Both molecules use the standardized 14-slot vocabulary of
\S\ref{sec:representation}: slots 1--4 hold the backbone atoms
$\mathrm{N},\mathrm{C}_\alpha,\mathrm{C},\mathrm{O}$, and slots 5--14
hold the side-chain heavy atoms in the fixed per-type order of
Table~\ref{tab:atoms} (at most ten; tryptophan is the only residue type
that uses all of them). When a residue provides fewer atoms than its
type allows, which is always the case for generated residues before
side-chain completion and for any unresolved atom in an observed
structure, the
unused slots are padded at the $\mathrm{C}_\alpha$ coordinate and
flagged as padding. Padded slots are excluded wherever atom slots are
enumerated: from the $k$-NN edge distances below, from the atom-level
softmax of Eq.~\ref{eq:atompool}, from the atom averages
$\bar{\mathbf{f}}$ and $\beta^{(e)}_j$, and from the gate indices
$\psi_p$ of Eq.~\ref{eq:coordupdate}.

\begin{table}[h]
\centering
\small
\caption{Side-chain heavy atoms occupying slots 5--14, in slot order;
slots 1--4 hold the backbone atoms.}
\label{tab:atoms}
\begin{tabular}{@{}ll@{\hspace{2.5em}}ll@{}}
\toprule
Res. & Slots 5--14 & Res. & Slots 5--14 \\
\midrule
Gly & --- & Ser & CB, OG \\
Ala & CB & Thr & CB, OG1, CG2 \\
Val & CB, CG1, CG2 & Cys & CB, SG \\
Leu & CB, CG, CD1, CD2 & Pro & CB, CG, CD \\
Ile & CB, CG1, CG2, CD1 & Phe & CB, CG, CD1, CD2, CE1, CE2, CZ \\
Asp & CB, CG, OD1, OD2 & Tyr & CB, CG, CD1, CD2, CE1, CE2, CZ, OH \\
Asn & CB, CG, OD1, ND2 & His & CB, CG, ND1, CD2, CE1, NE2 \\
Glu & CB, CG, CD, OE1, OE2 & Met & CB, CG, SD, CE \\
Gln & CB, CG, CD, OE1, NE2 & Trp & CB, CG, CD1, CD2, NE1, CE2, CE3, CZ2, CZ3, CH2 \\
Lys & CB, CG, CD, CE, NZ & Arg & CB, CG, CD, NE, CZ, NH1, NH2 \\
\bottomrule
\end{tabular}
\end{table}

\textbf{Edges.}
Context edges are constructed within each molecule and interface edges
between the two, both by $k$-NN over residues with $k{=}9$
(Table~\ref{tab:hyper}); the distance between two residues is the
minimum Euclidean distance over all pairs of non-padded atom slots, so a
single resolved contact suffices to rank a pair. Each interface edge
$e=(i,b_e)$ pairs an epitope residue $i$ with an antibody residue $b_e$
and carries the surface patch $\mathcal{V}_i$ of the epitope residue.

\subsection{Surface Patch Construction}
\label{app:surface}

The antigen surface is computed once per context with MSMS
\citep{sanner1996reduced} (1.5\,\AA{} probe radius); each vertex carries
the MSMS normal, oriented outward from the antigen. Following
\S\ref{sec:overview}, only vertices within $10\,\AA$ of any epitope atom
are retained. Each retained vertex is assigned to its nearest epitope
residue, which yields per-residue patches with $\approx$72 raw vertices
on average; each patch is randomly subsampled, or zero-padded and
masked, to a fixed size of $M{=}50$ vertices (Table~\ref{tab:hyper})
before batching. The mesh $\{\mathbf{v}_j\}$ is therefore part of the
fixed antigen context: it is precomputed before refinement, never
moves, and the same vertex assignment is reused at every decoding step.
Padded vertices are excluded from the vertex-level softmax of
Eq.~\ref{eq:surfattn} and from the attended centroid
$\bar{\mathbf{v}}_e$, so they can neither receive nor contribute
attention mass.

\subsection{Descriptor Composition and Attention Axes}
\label{app:descriptor}

Eq.~\ref{eq:geomfeat} is evaluated, per interface edge, on the grid of
valid atom slots $\times$ real patch vertices, that is, non-padded slots $p$ of
$a$ against non-padded vertices $j$ of $\mathcal{V}_i$, in the frame
of $a$; its 40 channels tally as $16+4+3+1+16$ across $\phi_r$,
$\mathbf{A}$, $\hat{\mathbf{u}}$, $c_{pj}$, and $\mathbf{a}_{p}$. The
cosine $c_{pj}$ is frame-independent by construction
(Eq.~\ref{eq:geomfeat}) and its sign is interpretable: $c_{pj}>0$ when
atom $p$ approaches vertex $j$ from the solvent side along the outward
normal, $c_{pj}<0$ when the approach direction points into the antigen.

The two attention levels of \S\ref{sec:interaction} normalize over the
two axes of this grid separately: the atom-level weights $\alpha_{pj}$
of Eq.~\ref{eq:atompool} softmax over the valid slots $p$ for each
vertex, and the surface-level weights $\beta^{(e)}_{pj}$ of
Eq.~\ref{eq:surfattn} softmax over the real vertices $j$ for each slot.
Masked entries are excluded from their normalization rather than
zeroed after it, so no probability mass leaks onto padding. The readout
$\mathbf{f}_p$, its valid-slot average $\bar{\mathbf{f}}$, the
normalized embedding $\mathbf{r}_e$, and the edge message
$\mathbf{m}_e$ (Eq.~\ref{eq:edgemessage}) then follow the main text;
all are strictly $\mathrm{SE}(3)$-invariant because every input is
frame-aligned (\S\ref{sec:representation}).

\subsection{Transient Coordinates and Layer-Wise Recurrence}
\label{app:recurrence}

Inside the interaction encoder, coordinates live in a \emph{local
complex}: the antibody residues together with the epitope residues of
their interface edges. At the start of each refinement round, every
coordinate in the local complex is re-imposed from the current
states, observed on the antigen side and predicted on the antibody
side; within the round, the encoder
maintains transient atomic states $\tilde{\mathbf{x}}_{i,p}$ on the
epitope side that evolve across layers through Eq.~\ref{eq:coordupdate},
while antibody-side coordinates enter as the current predictions. The
transient states exist only inside the round: they are re-initialized
at the next round, enter no loss, and never persist across decoding
steps, so the ground-truth antigen remains the fixed context everywhere
it is consumed.

Formally, let $\tilde{\mathbf{x}}^{(\ell)}$ collect the transient
coordinates entering encoder layer $\ell$ and $\mathbf{h}^{(\ell)}$ the
node states. Each layer interleaves two operators:
\begin{enumerate}
    \item an \emph{equivariant inter-chain update}: a multi-channel EGNN
    layer~\citep{satorras2021egnn} over the inter-chain edges whose radial (pair-distance)
    features are computed from the current coordinates of all local
    nodes, including the epitope-side states refined at earlier
    layers;
    \item the \emph{surface-attention update} of Eq.~\ref{eq:coordupdate},
    which refines the epitope-side states. Every quantity it consumes
    is recomputed from the states entering the layer: the residue
    frames $(\mathbf{R}_a,\mathbf{t}_a)$ (Eq.~\ref{eq:frame}), the
    descriptors $\mathbf{g}_{pj}$ and queries
    $\mathbf{q}_p^{\mathrm{loc}}$
    (Eqs.~\ref{eq:geomfeat}--\ref{eq:surfattn}), the atom-level weights
    $\alpha_{pj}$, and the attended centroids $\bar{\mathbf{v}}_e$.
\end{enumerate}
The composition realizes the recurrence
$\tilde{\mathbf{x}}^{(\ell+1)}=F_\ell(\tilde{\mathbf{x}}^{(\ell)},
\{\mathbf{v}_j\},\mathbf{h}^{(\ell)})$. The epitope-side states written
by operator~2 at layer $\ell$ are read by operator~1 at layers
$\ell{+}1,\ell{+}2,\dots$ exclusively through their pair-distance
features, which is how the update conditions downstream
distance contexts, whereas operator~2 itself consumes only the current
antibody coordinates and the fixed mesh. No layer therefore mixes
quantities defined in different geometric states.

Table~\ref{tab:provenance} summarizes where every state consumed by the
encoder comes from, how it evolves, and which training losses touch it.
Complex-structure prediction is the degenerate configuration with an
empty mask (the full antibody sequence is observed), and affinity
optimization adds Gaussian noise to the template initialization.

\begin{table}[h]
\centering
\small
\caption{Provenance of the coordinate and sequence states consumed by
the interaction encoder.}
\label{tab:provenance}
\begin{tabular}{@{}llll@{}}
\toprule
State & Initialization & Evolution & Training loss \\
\midrule
antigen atoms & observed context & fixed & --- \\
mesh $\{\mathbf{v}_j\}$, normals & precomputed (MSMS) & fixed & --- \\
\midrule
sequence (framework) & observed & fixed & --- \\
sequence (CDRs) & $[\mathrm{MASK}]$ & committed per round & $\mathcal{L}_{\mathrm{seq}}$ \\
coordinates (all) & framework template & refined per round & structure terms \\
transient states $\tilde{\mathbf{x}}_{i,p}$ & from observed antigen & Eq.~\ref{eq:coordupdate} only & none \\
shadow paratope & predicted & predicted per round & $\mathcal{L}_{\mathrm{SP}}$ \\
inter-chain edge distances & predicted & predicted per round & $\mathcal{L}_{\mathrm{ed}}$ \\
RMSD head & predicted & predicted per round & $\mathcal{L}_{\mathrm{pRMSD}}$ \\
\bottomrule
\end{tabular}
\end{table}

\subsection{SE(3)-Equivariance of the Interface Update}
\label{supp:geometric_properties}

\paragraph{Proof of $\mathrm{SE}(3)$-equivariance.}
Let $g=(\mathbf{R},\mathbf{t})\in\mathrm{SE}(3)$ act on \emph{all} input
point sets, atomic coordinates and the precomputed mesh alike, as
$\mathbf{x}'=\mathbf{R}\mathbf{x}+\mathbf{t}$ and
$\mathbf{v}'_j=\mathbf{R}\mathbf{v}_j+\mathbf{t}$.
\begin{enumerate}
    \item \emph{Frames.} The Gram--Schmidt construction of
    Eq.~\ref{eq:frame} is equivariant:
    $\mathbf{R}'_a=\mathbf{R}\mathbf{R}_a$ and
    $\mathbf{t}'_a=\mathbf{R}\mathbf{t}_a+\mathbf{t}$.
    \item \emph{Descriptors and queries.} Hence
    $\mathbf{d}'_{pj}=\mathbf{R}'^{\top}_a(\mathbf{x}'_{a,p}-\mathbf{v}'_j)
    =\mathbf{d}_{pj}$ and
    $\vec{n}^{\mathrm{loc}\prime}_j=\vec{n}^{\mathrm{loc}}_j$, so
    $\mathbf{g}_{pj}$, the atom-level weights $\alpha_{pj}$, the vertex
    summaries $\bar{\mathbf{g}}^{(e)}_j$, and the queries
    $\mathbf{q}_p^{\mathrm{loc}}$ are strictly invariant.
    \item \emph{Attention and gates.} Both arguments of the logit in
    Eq.~\ref{eq:surfattn} are invariant, so
    $\beta^{(e)\prime}_{pj}=\beta^{(e)}_{pj}$, and consequently the
    readout $\mathbf{f}_p$, its average $\bar{\mathbf{f}}$, the
    normalized embedding $\mathbf{r}_e$, the edge message
    $\mathbf{m}_e$, and the gates $\psi_p(\mathbf{m}_e)$ are invariant.
    \item \emph{Centroid and displacement basis.} The attended centroid
    co-transforms,
    $\bar{\mathbf{v}}'_e=\sum_j\beta^{(e)}_j(\mathbf{R}\mathbf{v}_j+\mathbf{t})
    =\mathbf{R}\bar{\mathbf{v}}_e+\mathbf{t}$, so translation cancels in
    the displacement basis:
    $(\mathbf{x}'_{a,p}-\bar{\mathbf{v}}'_e)
    =\mathbf{R}(\mathbf{x}_{a,p}-\bar{\mathbf{v}}_e)$.
    \item \emph{Update.} Therefore
    $\tilde{\mathbf{x}}^{\prime}_{i,p}
    =\mathbf{R}\tilde{\mathbf{x}}_{i,p}+\mathbf{t}
    +\mathbf{R}\,\Delta\tilde{\mathbf{x}}_{i,p}$: the update of
    Eq.~\ref{eq:coordupdate} transforms covariantly under $g$.
    \item \emph{Induction.} The inter-chain operator of step~1 is an
    EGNN layer built from pair distances (invariant radial features)
    and point differences (covariant directions) with invariant
    messages, and is likewise equivariant. Taking the co-transforming
    input point sets as the base case, equivariance composes through
    the full encoder by induction over layers.
\end{enumerate}

\subsection{Generation Procedure}
\label{app:generation}

Training and decoding share the same masked-prediction core. Each
training step draws a masking configuration (which determines the task,
\S\ref{sec:generation}), reveals a random, annealed fraction of the
masked residues as conditioning, initializes the masked sequence tokens
to $[\mathrm{MASK}]$ and the masked coordinates from the template, and
runs $R$ refinement rounds of the interface encoder with the node-state
memory carried across rounds; parameters are then updated under
Eq.~\ref{eq:total_loss}. Decoding mirrors this construction: starting
from the same masked initialization, each reveal round runs the $R$
refinement rounds with the memory carried over and commits tokens as in
\S\ref{sec:generation}; the antibody is then rigidly aligned to the
predicted paratope (Kabsch). Candidate selection at evaluation time is
fixed per task in \S\ref{sec:setup} (oracle-AAR best-of-five for the
design tasks; lowest predicted per-residue RMSD over ten draws for
structure prediction).

Because each committed token is drawn from
$\mathrm{softmax}(\mathbf{z}/\tau)$ and the commit order is randomized,
decoding defines a distribution over sequence--structure designs whose
concentration is controlled by $\tau$; $\tau\!\rightarrow\!0$ recovers
deterministic greedy decoding. Diversity is thus a property of the
decoding procedure by construction, and we measure it empirically in
Appendix~\ref{app:diversity}. The paragraph below makes the
correspondence with absorbing-state discrete diffusion
\citep{austin2021structured} precise.

\paragraph{Correspondence with absorbing-state discrete diffusion.}
Let $x=(x_1,\dots,x_n)$ be the target-region sequence over the
amino-acid vocabulary $\mathcal{V}$, extended with the absorbing state
$\mathrm{m}=[\mathrm{MASK}]$, and let $c$ collect the conditioning
context (antigen, framework, and already committed residues). Write
$x^{(s)}$ for the decoding state after reveal round $s$, with $x^{(0)}=
\mathrm{m}^{n}$. The decoder logits entering round $s$ are produced with
the recurrent node-state memory carried across rounds and the
coordinates refined so far; both are deterministic functions of the
committed tokens and the fixed context $c$. Conditioned on $c$ and this
decoder state, the one-step transition factorizes over positions:
still-masked positions follow Eq.~\ref{eq:reverse}, and committed
positions are never re-masked: revealed states are absorbing along the
chain. By telescoping, the schedule induces an exactly linear
unmasking trajectory,
\begin{equation}
\mathbb{E}\big[\gamma_s\big]
=\prod_{u=1}^{s}\left(1-\rho_u\right)
=\frac{K-s}{K},
\qquad s=0,\dots,K,
\label{app:eq:schedule}
\end{equation}
where $\gamma_s$ is the masked fraction, the discrete counterpart of
the linear schedule of absorbing-state diffusion
\citep{austin2021structured}. Three boundaries of the correspondence
follow. \textbf{Kernel:} the logits condition on the recurrent hidden
state and the refined coordinates as above, so the process is Markov in
the sequence only jointly with this decoder state; the correspondence
is therefore drawn at the level of the induced marginal unmasking
schedule of the sequence channel (Eq.~\ref{app:eq:schedule}), not as an
ELBO decomposition of the joint sequence--structure process.
\textbf{Training:} the masked-prediction objective of
\S\ref{sec:generation} lies in the simplified masked-cross-entropy
family used to train absorbing-state diffusion
\citep{sahoo2024masked}, with the corruption level swept by a
curriculum instead of sampled from a uniform prior; no ELBO identity is
claimed under this curriculum.
\textbf{Scope:} only the sequence carries absorbing-state semantics;
coordinates are initialized from a template (or from optimized noise,
for affinity optimization) and refined by the geometric updates of
\S\ref{sec:interaction}, not diffused.

\section{Training and Task Unification}
\label{app:impl}

\textbf{Surface construction} follows Appendix~\ref{app:surface}
(MSMS, per-residue patches subsampled to $M{=}50$).

\textbf{Training details.}
We train with Adam~\citep{kingma2015adam} and an exponentially decayed learning rate. Following
standard teacher-forcing annealing, each training step reveals a random
fraction of the masked residues as conditioning, where the fraction is
annealed from near zero toward larger values, so that the model learns to
decode from every intermediate state of the generative process. All
tasks are trained jointly in a single run: the masking configuration of
each step determines the task (\S\ref{sec:generation}), the same
objective and schedules apply to every configuration, and no task is
fine-tuned or adapted separately. The affinity-scoring pathway likewise
adds only a lightweight prediction head on the shared interface
representation; this head is the single component fitted outside the
unified run (see \textbf{Task unification}). All
schedules and loss weights are listed in Table~\ref{tab:hyper}; none is
tuned per task.

\textbf{Task unification.}
All tasks share a single architecture, a single interface operator set,
and the single training objective of Eq.~\ref{eq:total_loss}, evaluated
with one shared checkpoint; no task-specific adaptation is performed,
and within the shared generator no task-specific architectural
component or loss term exists. Which
terms of the objective are active at each step is determined entirely by
the masking configuration. In complex structure prediction the antibody
sequence is fully observed (no position is masked), so the masked-token
term $\mathcal{L}_{\mathrm{seq}}$ is vacuous by construction, and this
fully observed configuration corresponds to the conditioned endpoint of
the context annealing described above, the same spectrum of masking
states the model is trained on. In affinity optimization the same
generator is used frozen and unchanged; the $\Delta\Delta G$ regressor
is the single component fitted outside the shared model: a lightweight
prediction head on the shared interface representation, fit beforehand
on designed variants of the training complexes only
(Appendix~\ref{app:protocol}) and injected frozen as the
differentiable affinity proxy. Its supervision and usage are external
to the generator: ascent steps differentiate the frozen regressor
through the frozen generator with respect to the template-initialization
noise, updating neither the regressor nor the generator, and the
identical frozen regressor scores every compared method
(\S\ref{sec:affinity}), introducing no task-specific advantage. Task adaptation thus operates
entirely through masking, initialization, and sampling or optimization
configurations, consistent with \S\ref{sec:generation}.

\textbf{Loss terms and weights.}
Table~\ref{tab:losses} lists the terms of Eq.~\ref{eq:total_loss}.

\begin{table}[h]
\centering
\small
\caption{Loss terms of Eq.~\ref{eq:total_loss}; structure, docking, and
auxiliary terms are smooth-$\ell_1$ penalties.}
\label{tab:losses}
\begin{tabular}{@{}llcl@{}}
\toprule
Term & Definition (one line) & Weight & Source \\
\midrule
$\mathcal{L}_{\mathrm{seq}}$ & per-round CE, masked residues & 1 & standard \\
$\mathcal{L}_{x}$ & Kabsch-aligned coordinates & 1 & \citet{kong2023end} \\
$\mathcal{L}_{\mathrm{bond}}$ & backbone/side-chain bond lengths & 1 & \citet{kong2023end} \\
$\mathcal{L}_{\mathrm{SP}}$ & shadow-paratope coordinates & 1 & \citet{kong2023end} \\
$\mathcal{L}_{\mathrm{ed}}$ & predicted inter-edge distances & 1 & \citet{kong2023end} \\
$\mathcal{L}_{\mathrm{FAPE}}$ & frame-aligned point error, $d_{\max}$ clamp & 0.5 & \citet{jumper2021highly} \\
$\mathcal{L}_{\mathrm{torsion}}$ & dihedral/bond-angle cosines & 0.2 & \citet{kong2023end} \\
$\mathcal{L}_{\mathrm{pRMSD}}$ & per-residue RMSD prediction & 1 & ours (auxiliary) \\
\bottomrule
\end{tabular}
\end{table}

\section{Evaluation Details}
\label{app:protocol}

\textbf{Data.}
\label{app:data}
Both the training pool and the benchmark follow the official data
pipeline of \citet{kong2023end}. Antibody--antigen complexes are
obtained from SAbDab~\citep{dunbar2014sabdab} (snapshot of November 12, 2022) as
IMGT-renumbered~\citep{lefranc2003imgt} structures with paired heavy--light chains; entries
with mis-annotated chains or malformed structures are dropped during
cleaning, leaving 5{,}370 valid complexes, whose CDR boundaries are read
directly from the IMGT numbering. For the benchmark split, CDR-H3
sequences are clustered with MMseqs2 at 40\% sequence identity, and
every cluster containing a complex of the hold-out benchmark is removed
from the training and validation pools, so no training complex shares a
CDR-H3 cluster with any test complex; 10\% of the remaining complexes
are held out for validation. A conserved framework template is extracted
from the training structures for template initialization. The benchmark
itself is the 60-complex RAbD set \citep{adolf2018rabd}, likewise
IMGT-renumbered. We note that SAbDab and SKEMPI share complexes (e.g.,
\texttt{1a2y}, \texttt{2b2x} appear in both), which is why per-task
test sets cannot be retained under multi-task training
(\S\ref{sec:setup}).

\textbf{Baselines.}
All baselines are standardized under the unified protocol of
\S\ref{sec:setup}: same benchmark, same evaluator implementations, and
same per-task selection rules. For structure prediction and affinity
optimization, some baselines are our locally re-trained or
re-reproduced versions on RAbD, run with the identical inputs and
initialization as AbGaze.

\textbf{$\Delta\Delta G$ regressor.}
The affinity proxy is the simple prediction head on the shared interface
representation described above: its weights are fit beforehand on
designed variants of the training complexes only (no test complex is
involved) and are frozen during optimization; the identical frozen
regressor scores every method.

\textbf{Metrics.}
DockQ is computed with the reference implementation of
\citet{basu2016dockq}
(\texttt{github.com/bjornwallner/DockQ}), applied to the
CDR-H3--antigen interface consistent with \citet{jin2022antibody}; RMSD
is the C$_\alpha$ RMSD of the full antibody (heavy and light chains) after
Kabsch alignment, with per-CDR RMSDs aligned per loop;
CAAR restricts AAR to binding residues within 6.6\,\AA{} of the epitope.
Experiments run on two NVIDIA A800 GPUs (80\,GB each).

\section{Comparison with IgGM}
\label{app:iggm}

IgGM~\citep{wang2025iggm} is a generative model for functional antibody
design that reports CDR-level AAR on its own SAbDab split. We compare
against it in Table~\ref{tab:iggm} rather than in the main experiments,
for three reasons. First, the test sets differ: IgGM evaluates on its own
SAbDab split (post-2023 entries), while our unified protocol evaluates
all tasks on RAbD (\S\ref{sec:setup}); placing IgGM in the main table
would require either re-training on a different split or comparing
numbers across incompatible test sets. Second, our training set is
smaller: IgGM trains on a larger set of antibody--antigen complexes
collected with additional filtering, so AbGaze is disadvantaged in this
comparison. Third, IgGM models backbone atoms only, which precludes
comparison on full-atom metrics such as side-chain lDDT or
atom-level interface quality.

Despite these handicaps, AbGaze outperforms IgGM on five of six CDRs,
with the largest margins on the binding-critical loops H2 (+5.0 points)
and L3 (+4.2 points); IgGM leads only on L1 (+2.2 points).

\begin{table}[htbp]
  \centering
  \small
  \caption{CDR-level AAR comparison with IgGM on its own SAbDab split
  (post-2023); IgGM numbers are from its published Table~2.}
  \label{tab:iggm}
  \begin{tabular}{@{}cccc@{}}
    \toprule
    CDR & AbGaze & IgGM & AbGaze wins \\
    \midrule
    H1 & \textbf{0.752} & 0.740 & \checkmark \\
    H2 & \textbf{0.694} & 0.644 & \checkmark \\
    H3 & \textbf{0.397} & 0.360 & \checkmark \\
    L1 & 0.728 & \textbf{0.750} & $\times$ \\
    L2 & \textbf{0.753} & 0.743 & \checkmark \\
    L3 & \textbf{0.677} & 0.635 & \checkmark \\
    \bottomrule
  \end{tabular}
\end{table}

\section{Diversity Analysis}
\label{app:diversity}

As stated in Appendix~\ref{app:generation}, decoding defines a
distribution over designs whose concentration is controlled by $\tau$.
We quantify this on the all-CDR design setting (five samples per
target, $\tau=0.5$): Table~\ref{tab:diversity} reports, per CDR, the
number of distinct sequences among the five samples and the mean
pairwise normalized Hamming distance between them.

\begin{table}[h]
\centering
\small
\caption{Design diversity on all-CDR design (five samples per target,
$\tau=0.5$).}
\label{tab:diversity}
\begin{tabular}{@{}lccc@{}}
\toprule
CDR & Distinct ($\uparrow$) & Diversity ($\uparrow$) & \% all-same \\
\midrule
H1 & 2.60 & 0.092 & 21.7\% \\
H2 & 2.93 & 0.127 & 16.7\% \\
H3 & \textbf{4.92} & \textbf{0.458} & \textbf{1.7\%} \\
L1 & 2.00 & 0.070 & 50.0\% \\
L2 & 1.42 & 0.062 & 70.0\% \\
L3 & 3.93 & 0.181 & 3.3\% \\
\midrule
All & 2.97 & 0.165 & 27.2\% \\
\bottomrule
\end{tabular}
\end{table}

Diversity differs strongly across loops: CDR-H3, the loop that
dominates binding specificity and tolerates the most sequence
variation, is nearly saturated (4.92/5 distinct, pairwise Hamming
0.458, and only 1.7\% of targets produce identical H3 across all five
samples), while the framework-proximal L2 is the most conservative
(1.42 distinct, 70\% identical). This pattern is consistent with the
greater sequence tolerance of H3 relative to the framework-proximal
loops.

\section{Gains without Diverse Sampling}
\label{app:gains_without_sampling}

The improvements reported in the main experiments do not rely on sampling
diverse candidates. To show this, we replace the diverse decoding of
Appendix~\ref{app:generation} with a single deterministic
regression-style design: all masked positions of the target region are
predicted and committed in one forward pass from the fully masked state,
without the iterative commit-and-resample schedule or the temperature
$\tau$. Table~\ref{tab:regression_allcdr} evaluates this single-draw
instantiation on all-CDR design under otherwise identical conditions
(RAbD, same evaluator), against the full generative results of AbFlow
and dyMEAN from Table~\ref{tab:main_results}. AAR/CAAR are pooled over CDR
residues, and RMSD is the C$_\alpha$ RMSD of the full antibody (heavy
and light chains) after Kabsch alignment.

\begin{table}[h]
\centering
\small
\caption{All-CDR design on RAbD without diverse sampling; notation as
in Table~\ref{tab:main_results}.}
\label{tab:regression_allcdr}
\begin{tabular}{l *{6}{c}}
\toprule
\textbf{Method}
& \textbf{AAR}$\uparrow$ & \textbf{CAAR}$\uparrow$ & \textbf{RMSD}$\downarrow$
& \textbf{DockQ}$\uparrow$ & \textbf{lDDT}$\uparrow$ & \textbf{TM-Score}$\uparrow$ \\
\midrule
dyMEAN & 60.1\% & 50.3\% & 1.357 & 0.396 & 0.803 & 0.965 \\
AbFlow & 59.7\% & 49.8\% & 1.104 & 0.379 & 0.815 & 0.971 \\
\midrule
\textbf{AbGaze (Reg.)} & 64.0\% & 54.9\% & \textbf{1.021} & 0.407 & \textbf{0.834} & \textbf{0.974} \\
AbGaze (Full) & \textbf{66.2\%} & \textbf{57.1\%} & 1.052 & \textbf{0.422} & 0.831 & 0.973 \\
\bottomrule
\end{tabular}
\end{table}

Switching to the single deterministic design has little cost. Only
AAR drops slightly, from 66.2\% to 64.0\%, and its contact-restricted
variant likewise, while still leading the stronger baseline by 3.9
points; every other metric is essentially unchanged: mean RMSD improves
(1.052 to 1.021\,\AA{}), lDDT and TM-Score are comparable, and
DockQ shifts from 0.422 to 0.407, above both baselines
(0.396/0.379). A single deterministic regression-style design therefore
already outperforms the full generative models of AbFlow and dyMEAN on
every metric: the advantage stems from the orientation-aware interface
representation itself, not from diverse sampling.

We nevertheless retain the diverse absorbing-state decoding in the main
protocol, for reasons orthogonal to benchmark performance. First, parity:
every compared method is sampled and selected identically
(\S\ref{sec:setup}), so the stochastic protocol is the like-for-like
configuration. Second, practice: antibody design workflows consume a
\emph{pool} of diverse candidates for screening, filtering, and
best-of-$N$ retrieval, whereas a deterministic regressor collapses the
pool to a single mode per target; diversity concentrates on the loops
with the most design freedom (Appendix~\ref{app:diversity}).
Diverse generative decoding is thus adopted because it matches the actual
requirements of the design task.

\section{Hyperparameters}
\label{app:hyper}

Table~\ref{tab:hyper} lists all hyperparameters; none is tuned per task.

\begin{table}[h]
\centering
\small
\caption{Hyperparameters.}
\label{tab:hyper}
\setlength{\tabcolsep}{4pt}
\begin{tabular}{@{}lll@{}}
\toprule
Hyperparameter & Value & Description \\
\midrule
embed dim & 64 & residue/atom embedding \\
hidden size & 128 & message-passing hidden size \\
encoder layers & 3 & equivariant encoder layers \\
refinement rounds $R$ & 3 & rounds per decoding step \\
$k$ (neighbors) & 9 & KNN neighbors per residue \\
atom channels $C$ & 14 & 4 backbone + $\le$10 side-chain atoms \\
surface size $M$ & 50 & vertices per epitope residue \\
RBF bases / cutoff & 16 / 10\,\AA & distance embedding \\
$\lambda_{\mathrm{FAPE}}$ / $\lambda_{\mathrm{torsion}}$ & 0.5 / 0.2 & loss weights \\
$d_{\max}$ & 10\,\AA & FAPE clamping \\
reveal rounds $K$ & 9 & unmasking steps ($T{=}K{+}1{=}10$) \\
temperature $\tau$ & 0.5 & candidate diversity \\
prediction retries & 10 & draws ranked by predicted RMSD \\
affinity candidates & 30 & optimized designs per target \\
optimization steps / lr & 5 / 1.0 & ascent on template noise \\
optimizer / lr & Adam / $10^{-3}\!\rightarrow\!10^{-4}$ & exponential decay \\
gradient clipping & 1.0 & max global norm \\
sequence-loss warmup & 10 epochs & linear ramp of $\lambda_{\mathrm{seq}}$ \\
batch size / epochs & 16 / 500 & \\
hardware & 2$\times$ NVIDIA A800 (80\,GB) & \\
\bottomrule
\end{tabular}
\end{table}

\end{document}